\def\year{2022}\relax
\documentclass[letterpaper]{article} 
\usepackage{aaai22}  
\usepackage{times}  
\usepackage{helvet}  
\usepackage{courier}  
\usepackage[hyphens]{url}  
\usepackage{graphicx} 
\usepackage{natbib}  
\usepackage{caption} 
\DeclareCaptionStyle{ruled}{labelfont=normalfont,labelsep=colon,strut=off} 
\usepackage{algorithm}
\usepackage{algorithmic}
\usepackage{lipsum}
\usepackage{mathtools}
\usepackage{cuted}
\usepackage{amsfonts}

\usepackage{newfloat}
\usepackage{listings}
\floatstyle{ruled}
\newfloat{listing}{tb}{lst}{}
\floatname{listing}{Listing}
\title{More Data Can Lead Us Astray: Active Data Acquisition\\ in the Presence of Label Bias}

\title{More Data Can Lead Us Astray: Active Data Acquisition \\ in the Presence of Label Bias}
\author {
    Yunyi Li,
    Maria De-Arteaga, 
    Maytal Saar-Tsechansky 
}
\affiliations {
    University of Texas at Austin\\
    yunyi.li, dearteaga, maytal.saar-tsechansky @mccombs.utexas.edu
}

\usepackage{bibentry}

\begin{document}

\maketitle

\begin{abstract}
An increased awareness concerning risks of algorithmic bias has driven a surge of efforts around bias mitigation strategies. A vast majority of the proposed approaches fall under one of two categories: (1) imposing algorithmic fairness constraints on predictive models, and (2) collecting additional training samples. Most recently and at the intersection of these two categories, methods that propose active learning under fairness constraints have been developed. However, proposed bias mitigation strategies typically overlook the bias presented in the observed labels. In this work, we study fairness considerations of active data collection strategies in the presence of label bias. We first present an overview of different types of label bias in the context of supervised learning systems. We then empirically show that, when overlooking label bias, collecting more data can aggravate bias, and imposing fairness constraints that rely on the observed labels in the data collection process may not address the problem. Our results illustrate the unintended consequences of deploying a model that attempts to mitigate a single type of bias while neglecting others, emphasizing the importance of explicitly differentiating between the types of bias that fairness-aware algorithms aim to address, and highlighting the risks of neglecting label bias during data collection. 
\end{abstract}

\section{Introduction}
There is sufficient understanding that machine learning (ML) algorithms can easily replicate and even exacerbate societal biases. Abundant empirical evidence of biased ML systems has been found in a variety of  domains, including criminal justice~\cite{machinebias}, healthcare~\cite{obermeyer2019dissecting}, human resources~\cite{pessach2020algorithmic}, and content moderation~\cite{sap2019risk,davidson2017automated}. In many of these domains, algorithms are making or supporting high-stakes, life-changing decisions. Awareness of the risks of biases in ML systems has led to an exploration of methods to mitigate those biases. A majority of the works that propose approaches to mitigate algorithmic bias measures bias at a group level~\cite{mitchell2018prediction, verma2018fairness} and have been focused on one of two categories. One line of work aims to use available data to train algorithms that yields better group fairness measures. This is typically achieved by imposing different types of fairness constraints~\cite{bellamy2018ai, d2017conscientious}, including pre-processing data transformations~\citep{ zemel2013learning, louizos2015variational, lum2016statistical, adler2018auditing, turchetta2016safe, del2018obtaining}; in-processing optimization constraints~\citep{ woodworth2017learning, zafar2017fairness, agarwal2018reductions, russell2017worlds}; and post-processing group-specific classification thresholds~\citep{feldman2015computational, hardt2016equality}. The second approach aims to mitigate bias via additional data collection, which may be guided by a number of criteria, including the desire to cost-effectively produce a distribution that is expected to yield better generalization performance and smaller biases. For example, to prevent harms caused by ML systems, there have been numerous calls to obtain training datasets that are more inclusive, diverse and representative of the populations of interest~\cite{chen2018my, fazelpour2022diversity, gebru2021datasheets, mitchell2019model, veale2017fairer, holstein2019improving}, including calls by policy makers~\cite{commissie2021proposal,ai2019high}.


Because data collection--and especially label acquisition--is very costly, active learning is often used to assist the data collection process, so as to cost-effectively acquire labels that are particularly beneficial for learning ~\cite{saar2004active, geva-tsechansky-alp}. In contrast with traditional supervised learning, (pool-based) active learning is a framework in which the learner's goal is to proactively select a subset of examples to be labeled from a pool of unlabeled instances. Typically, an active learner is initially trained on a small set of labeled examples and adaptively determines the batch of unlabeled instances that would be most advantageous to be labeled next. Most active learning strategies choose instances to be labeled based on some notion of uncertainty, such as entropy of predictions~\cite{shannon1948mathematical}. However, an instance's utility score (how beneficial it would be to include it in the training data) can be designed to depend on a number of criteria, and different heuristics can be used to achieve different performance needs. As recent work has noted, these heuristics can be modified to reflect fairness criteria. For example, if the goal is to learn a model that can mitigate the performance disparity across groups, then the active learning utility function can be designed to incorporate this objective \citep{abernethy2020adaptive}. 
Thus, active learning's suitability for efficient and purpose-driven data collection has spurred interest at the intersection of algorithmic fairness and efficient data collection, and algorithms that propose active learning under fairness constraints have been developed. For instance, \citet{abernethy2020adaptive} proposed adaptive sampling to reduce disparate performance by considering group membership when estimating the probability of sampling an instance to be labeled. \citet{anahideh2022fair} proposed a method called ``fair active learning'' (FAL), which includes fairness improvements in the utility function in addition to overall model performance (eg. overall model accuracy). 

Approaches for mitigating bias by imposing fairness constraints and existing approaches for fairness-oriented active data collection share one important commonality: they all assume (albeit often implicitly) that the observed label can be considered to be an unbiased ``gold standard". Thus, the proposed bias mitigation strategies typically overlook bias present in the observed labels. This can be a problem because historical prejudice, bias and unequal access to opportunities do not only affect covariates, but may also affect the labels used for training across domains. For example, \citet{obermeyer2019dissecting} found that one widely used algorithm in the U.S. healthcare system meant to identify patients that will benefit most from enrollment in ``high-risk care management'' programs exhibited bias against Black patients due to label bias. The algorithm relied on healthcare costs as a proxy for health needs, but unequal access to care has historically lead to lower healthcare spending for Black patients as compared to white patients, resulting in the algorithm underestimating their needs. Racial bias in data labels has also been shown to be present in multiple widely used Twitter corpora annotated for offensive language, where tweets inferred to be African American English are more likely to be annotated as offensive language ~\cite{sap2019risk}. Research has also shown that models trained on these corpora have the potential to reinforce the racial bias and further marginalize voice from minorities~\cite{sap2019risk}.  

In this work, we provide a structured overview of different types of label bias and empirically study fairness implications of active data collection algorithms in the presence of label bias. Our work is grounded on a disconnect between the characteristics of common active learning settings, and the underlying assumptions of proposed bias mitigation strategies. 
Overlooking label bias in active data collection may be particularly problematic given that some of the most common active learning applications rely on human-generated labels, such as crowdsourced annotations ~\cite{yan2011active}, which are highly bias prone due to human cognitive bias ~\cite{haselton2015evolution}. Furthermore, the settings used to motivate active learning with fairness constraints are often some in which label bias is a high risk. For instance, a recently proposed methodology termed ``fair active learning" (FAL)~\citep{anahideh2022fair} uses recidivism prediction as one of two motivating examples and evaluates the proposed approach in the COMPAS dataset, while recent work has highlighted the risk of label bias when using rearrest as a proxy for recidivism~\citep{bao2021s,fogliato2021validity}.

We study the impact of overlooking label bias during data collection empirically, using both simulations and real data. 
Our results show that if we overlook label bias while acquiring labels: 1) collecting more data can lead to exacerbated bias; 2) data-driven strategies to identify the ``disadvantaged group" based on performance gaps can lead to misidentification; 3) relative comparisons of bias based on performance gaps across models can be misleading, which may misguide model selection. 

In the next section, we briefly review extant research on algorithmic fairness (bias mitigation algorithms and calls for data collection), active learning, fairness-aware active learning, and label bias. Then we conceptualize label bias in the context of supervised learning systems and provide a structured overview of difference types of label bias. We then describe our methodology for assessing fairness performance of recently proposed fairness-aware active learning and most commonly used active learning strategies in the presence of label bias. Thereafter, we present the results and show three \emph{more data can lead us astray} patterns. We conclude by discussing the implications of our findings and opportunities for future research directions. 

\section{Related Work}\label{literature}
In recent years, a large body of research has focused on the development of \textbf{bias mitigation algorithms}. Most of them fall into one of three streams: pre-processing, in-processing, or post-processing. Pre-processing methods try to remove information related to sensitive feature(s).  The idea is that we first drop the sensitive feature(s), then learn a new feature space that removes the information correlated to the sensitive feature~\citep{zemel2013learning, louizos2015variational, adler2018auditing, calmon2017optimized}. Alternatively, in-process methods try to modify the loss function in order to penalized the disparities in performance, for example, adding a constraint or a regularization term to the existing loss function~\citep{zafar2017fairness, gordaliza2019obtaining, agarwal2018reductions, zafar2017fairness, woodworth2017learning, calders2009building}. 
Generally, the constraint is a quantitative operationalization of an underlying notion of fairness, and can be added to the objective function of any supervised machine leaning model. The trade off between the existing objective (eg. accuracy) and fairness metrics can be treated as a user parameter and thus can be adjusted based on different contexts and stakeholders’ interests. Post-processing methods try to find appropriate thresholds using the original scoring function for each group~\citep{hardt2016equality}. As they are using the original scoring function, there is no need to retrain the model, thus is preferred if computation expense is an issue, but may come at a cost of other notions of fairness~\citep{cheng2021social}. All of the aforementioned fairness-aware algorithms discuss the bias mitigation techniques with respect to observed labels, however, we consider label bias and differentiate observed labels from the labels of interest for algorithm training. 


Despite these ``fair" algorithm developments on fixed datasets, many researchers attribute the disparities to the unrepresentative of training samples~\cite{buolamwini2018gender, machinebias}, and advocate for \textbf{data collection} to avoid discrimination~\cite{chen2018my, gebru2021datasheets, mitchell2019model, veale2017fairer, holstein2019improving}. As we have mentioned in the introduction section, \textbf{active learning} approaches \citep{lewis1994sequential, settles2008analysis, settles2009active, zhu2009active, huang2010active} are considered a cost-effective way to acquire additional training instances, since the algorithm can guide the selection of the most informative set of instances to be labeled and added to the training set. Driven by the need of collecting more data to mitigate bias cost effectively, \textbf{fairness-aware active learning} methods have been proposed recently. \citet{anahideh2022fair} proposed a query strategy that sample the next batch of instances considering both overall model information gain and fairness measure. At each iteration, the instance(s) with maximum Shannon entropy~\cite{shannon1948mathematical} and expected fairness improvement would be labeled and added to the training set. However, the obtained labels are still assumed to be the ``gold standard''. Similarity, other very recent works also incorporate the fairness notion in active learning strategy design \citep{abernethy2020adaptive, sharaf2020promoting, cai2022adaptive}. All these new approaches inherit from classical active learning the assumption that the acquired label is a perfect match with the label of interest, but this assumption does not hold in a wide range of practical scenarios. 

A separate body of work has devoted significant attention to the presence of \textbf{bias in human-generated labels}. With the flourishing of crowdsourcing services~\cite{howe2008crowdsourcing, yan2011active}, such as Amazon Mechanical Turk, the data labeling process is increasingly reliant on crowd work~\citep{gray2019ghost}. Crowd workers can perform as well as domain experts in certain tasks ~\cite{snow2008cheap}, especially when the composition of the workers' pool is carefully curated in a task-dependant manner~\citep{allen2021scaling}. However, researchers have brought attention to the risk of annotators' cognitive bias~\cite{eickhoff2018cognitive, draws2021checklist}, stereotyping encoded in annotators' assessments~\cite{otterbacher2015crowdsourcing}, and uneven representations of demographic characteristics among annotators~\cite{barbosa2019rehumanized}. 
A number of factors, including task and instructions clarity ~\cite{wu2017confusing}, task design ~\cite{kazai2011crowdsourcing}, incentives ~\cite{shah2015double}, and quality control mechanisms ~\cite{ipeirotis2010quality, mcdonnell2016relevant}, have been demonstrated to affect the quality of the annotations ~\cite{draws2021checklist}. Furthermore, even when labels are collected from domain experts, this does not mean they are free of bias. For instance, in the context of healthcare, the quality of medical diagnoses and treatments in acute, cancer, and palliative pain care was compromised due to medical care providers' biases \citep{hoffman2016racial}. 



It is worth mentioning that there is a large body of literature that relaxes the assumption of perfect labels, focused on \textbf{learning from noisy labelers}. Researchers have proposed aggregating multiple noisy labelers' opinion either through majority voting \citep{zhang2016learning}, as well as learning probabilistic models to jointly estimate labelers' quality and gold standard labels~\cite{snow2008cheap, smyth1995inferring, dawid1979maximum, whitehill2009whose, rodrigues2013learning, welinder2010multidimensional, jin2002learning, liu2012variational, yan2010modeling}, or other heuristics \cite{huang2017cost, gao2020cost}.  These proposed algorithms, nevertheless, assume constrained forms of noise such as random noise, which exclude shared societal biases, or dismiss fairness considerations in their model evaluation. Recent work has aimed to move beyond these assumptions by leveraging disagreement and recognizing annotators' unique perspectives~\citep{davani2022dealing}, but there is still a lack of active learning approaches that tackle this problem.

While label noise in ML systems has received extensive attention, there are relatively few works explicitly focused on \textbf{label bias}. \citet{fogliato2020fairness} found that even small biases in observed labels is sufficient to lead to disparities of recidivism risk assessment tool's disparate performance on different racial groups. The fact that ``re-arrest''---the target variable used for training recidivism risk assessment tools---, is a different construct than ``re-offend''---the outcome the risk assessment tool aims to predict---can cause fairness issues; ``a model that appears to be fair with respect to rearrest could be an unfair predictor of re-offense''~\cite{fogliato2020fairness}. Label bias has been identified as a potential problem in other contexts such as child maltreatment hotline screenings~\cite{de2021leveraging}, healthcare~\cite{obermeyer2019dissecting}, and offensive language detection~\cite{sap2019risk}. \citet{obermeyer2019dissecting} and \citet{sap2019risk} attribute the performance disparity of a ML system to a specific label bias under the context they examined, and \citet{de2021leveraging} propose methodology that combines observed outcomes and human decisions to better approximate a construct of interest. Finally, the context dependent and complex relationship between the observed labels used for training ML models, the construct of interest that an algorithm aims to predict, and the decision making space has been discussed ~\cite{friedler2016possibility, passi2019problem,jacobs2021measurement}.


\section{How Bias May Creep into Labels} \label{label bias}
In this section, we describe how bias may creep into instances' labels used to train ML  models. Before we dissect different sources of label bias, let us define what we mean when we use the term \emph{label bias}. Given an observed label $\tilde Y $ that is usually readily accessible in a set of training examples, a construct of interest $Y^*$ that is ideal label for training, and a variable denoting a binary group membership $G$, we refer to label bias as a systematic mismatch between the construct of interest and the observed label, such that the relationship underlying the mismatch differs across groups. $\tilde Y$ exhibits label bias with respect to $Y^*$ if there is a group of relevance $g_0\in G$ such that,

\begin{equation}
P(Y^* =\tilde Y | G = g_0 ) \neq P(Y^* = \tilde Y | G \neq g_0 )
\end{equation}

For example, $G$ may correspond to gender and $g_0$ to women. Label bias differs from label errors given its non-random nature and tight relationship with group fairness. It is also important to note that according to this definition, the presence of label bias cannot be determined on the basis of data alone; it is defined in reference to a construct of interest $Y^*$ for training a supervised learning system, and thus depends on the predictive task that a supervised learning system is attempting to perform. Such bias may be a result of different phenomena, which we detail below.  

\textbf{Construct Gap} occurs when there is a mismatch in theoretical definitions between the construct of interest and the construct of observed label~\citep{jacobs2021measurement}. The reliance on a construct that does not match the construct of interest is typically motivated by the relative accessibility of one over the other. In some cases, this may occur because ML systems are often trained by repurposing previously collected datasets stored in organizational information systems. As a result, the goals during data collection may not match the goals during model development. For example, in the context of healthcare, financial incentives often result in detailed and meticulous data of insurance claims, which is then repurposed for multiple tasks in ML for healthcare.  
The risks of relying on insurance claims and spending information as a proxy for healthcare conditions are illustrated in a study by \citet{obermeyer2019dissecting} analyzing racial disparities in an algorithm used to prioritize patients that may benefit from care management programs meant for patients with complex health needs. 
The algorithm used health costs as a proxy label for health needs. Due to historical inequities, in the US the costs incurred by Black patients are often considerably lower than those of white patients with similar healthcare needs. As a result, the algorithm prioritized healthier white patients over more ill Black patients. 
Construct validity issues may also arise because high-level, complex objectives are often not directly quantifiable~\citep{passi2019problem}. As a result, it is necessary to choose proxies. Consider the task of predicting a student's likelihood to be successful in a college admission setting. It is impossible to fully capture the outcome of a `successful student' using one single measurable feature due to the complex and potentially contentious definition of `success'. In cases like this, a simplified construct such as `GPA' or `class ranking' is used, which may ignore various indicators of success, and inadvertently favor some population groups \citep{suresh2019framework}. 

\textbf{Label Measurement Bias} occurs when the construct we are interested in when training a supervised learning model is fully aligned with the construct we intended to measure, but measurement errors vary across groups~\citep{jacobs2021measurement}. While this may also be thought of as a proxy problem, label measurement bias differs from construct gap in that there is no \emph{conceptual} mismatch between the observed construct and the construct of interest. For example, measurement error from pulse oximetry is more prevalent for people with darker skin pigmentation. Compared to white patients, Black patients had roughly three times the rate of occult hypoxemia that was not identified by pulse oximetry \citep{sjoding2020racial}. Thus, if a ML model is trained to predict oxygen saturation \citep{ghazal2019using}, the label we observe does match the construct we are interested in, but the measurement error is higher for Black patients as compare to white patients. In the context of ongoing efforts to use ML models to automatically adjust ventilation settings \citep{ghazal2019using}, this label measurement bias could result in mis-adjustment or delayed adjustment of ventilator settings for Black patients. Given that measurement often relies on various technologies as well as sociotechnical processes, disparities in measurement error are common across domains.  

\textbf{Human Labeling Bias} arises in contexts when labels used for training ML models rely on human assessments, which is very common in the crowdsourced data collections. In many tasks, ranging from radiology applications to misinformation detection, there is frequently no unique gold standard~\citep{adamson2019machine,neumann2022justice}, and the construct of interest $Y^*$ may not be directly observable and may be subject to disagreement ~\cite{aroyo2019crowdsourcing}. For example, when assessing candidates for job applications, committee members may disagree with each other on whether someone is qualified for the job. In this case, the labeling disagreements can be rooted in the differing subjectiveties and value systems ~\cite{bless2014social}. Crowd workers, whose life experience brings important perspectives to certain tasks may be dramatically underrepresented in generic crowdsourcing platforms such as Mturk, and the voice from the already unrepresentative group can be further marginalized~\cite{davani2022dealing}. 
In addition to capturing a variety of perspectives, human-generated labels frequently 
capture societal biases and prejudices. Human labeling bias occurs when there is a mismatch between a construct of interest $Y^*$ and the labels provided by human labelers $\tilde Y$, such that a group is systematically disfavoured. 
For example,~\cite{otterbacher2018investigating} found that people who are more sexist, as measured by the Ambivalent Sexism Inventory, are less likely to recognize and report gender biases in image search results, thereby reinforcing social stereotypes. 

It is worth mentioning that these three types of label bias are not mutually exclusive. 
 For example, construct gap and human labeling bias may often co-occur, as the proxy chosen may correspond to human assessments. For instance, when using income or promotions as a proxy for job skills or potential, there is both a construct gap issue and a risk of human labeling bias, as previous promotion and raise decisions are made by managers. However, conceptually distinguishing between the different types of label bias can help reason about the pathways through which these may capture societal injustices, and the ways in which the relationship between $Y$ and $Y^*$ may vary across groups.   

\section{Methodology}
To empirically study the impact of overlooking label bias during a data collection process under (fairness-aware) active learning strategies, we use two datasets: the UCI Adult dataset~\cite{Dua:2019}, and an offensive language dataset~\cite{keswani2021towards}. In this section, we provide a detailed description of the five (fairness-aware) active learning algorithms we empirically evaluated, the two datasets, and the experiment settings.   


\subsection{Algorithms Evaluated} \label{algorithms}
We consider five active learning strategies: 1) Fair Active Learning; 2) Adaptive sampling; 3) Adaptive Sampling with uncertainty criteria; 4) Uncertainty Sampling; and 5) Random Sampling.  The first three strategies are bias-mitigation active learning algorithms, the forth strategy is a traditional and widely used active learning algorithm, and the last strategy is a naive sampling policy that is usually used as baseline for the development of active learning algorithms. All of the five algorithms iteratively select a batch of unlabeled instances, query labels for them, and add them to the training data, but use different heuristics for the instance selection. Below we introduce the five different heuristics:

\begin{itemize}
    \item \textit{Fair active learning (FAL)} introduces group fairness constraints to traditional active learning objectives. It selects instances to be labeled based on the linear combination of two criteria: uncertainty based Shannon entropy, and expected fairness improvement, measured using a group fairness metric \citep{anahideh2022fair}. The penalty parameter controlling the trade off between those two criteria is a user input. Empirically, \citet{anahideh2022fair} shows a considerable decrease in disparity while preserving accuracy when evaluating with respect to observed labels. 
    \item \textit{Adaptive sampling (Adaptive)} reduces disparate performance by either randomly selecting instances to be labeled from the disadvantaged group at a probability $p$, or randomly sampling instances from the full unlabeled pool at the probability $(1-p)$ \citep{abernethy2020adaptive}. The trade off parameter $p$ is a hyper-parameter, and the group considered the ``disadvantaged group" is determined in a data-driven way at each iteration, defined as the group for which the algorithm has a lower performance at a given point.
    \item \textit{Adaptive sampling with uncertainty criterion (Adapt. Uncert.)} is a natural extension to the original adaptive sampling method \citep{abernethy2020adaptive}. We implement this variant by adding an uncertainty criterion \citep{settles2009active} to the sampling process. After deciding which group to sample from, instead of randomly sampling an instance, we use Shannon entropy \citep{shannon1948mathematical} to select the instance that the algorithm is most uncertain about. This constitutes an active learning variant of adaptive sampling. Specifically, with probability $p$, we apply uncertainty sampling constrained to the ``disadvantaged group", and with probability $(1-p)$ we apply uncertainty sampling to the full unlabeled pool of instances. 
    \item \textit{Uncertainty sampling (Uncertainty)} chooses the instances in each run to be labeled based on which instances the current model is most uncertain about \citep{settles2009active}. A general method to measure uncertainty is Shannon entropy \citep{shannon1948mathematical}. 
    \item \textit{Random sampling (Random)} is commonly used as a baseline in the active learning literature. The method randomly selects an instance from the unlabeled pool. 
\end{itemize}

\subsection{Datasets}

\subsubsection{UCI Adult Dataset}\label{data:Adult}
We use the Adult dataset \citep{Dua:2019} from UCI Machine Learning Repository for income level prediction and we simulate a construct of interest for the dataset. The Adult dataset \citep{Dua:2019}, also known as "census income" dataset, is extracted from the 1994 census data in the United States, and is widely used for ML modeling and algorithmic fairness research. This dataset is a good example to represent construct gap bias, as the the labels in the dataset---whether a given adult earns more than \$50K per year---reflect historical inequalities that have resulted in lower wages for women and unpaid domestic labor. This in itself is not a problem with the data. But if income is assumed to be a proxy for other constructs of interest, such as contribution to the economy or deserving income based on skills, the construct gap can result in algorithmic bias. Labels in the dataset reveal dramatic economic disparities between men and women. Approximately one-third of men are reported to earn more than \$50K per year, while only one-tenth of women are reported to have same level of income. We use this dataset to conduct a first set of semi-synthetic experiments in which we simulate the relationship between $\tilde Y$ and $Y^*$, allowing us to have full control over this relationship in order to explore how label bias may affect active learning. For the simulation, we borrow the idea from \citet{jiang2020identifying}'s mathematical work and increased female positive rate so that it matches male positive rate. Specifically, we create a construct of interest, $Y^*$, by uniformly drawing a certain percentage of negative female instances and changing the label to be positive. This results in a label that satisfies statistical parity in the data, having the same proportion of each group labeled as positive.

\subsubsection{Offensive Language Dataset} \label{data:hate_speech}
Hate speech and offensive language identification is difficult because what is considered offensive depends on the social context. Derogatory terms towards African American communities have been reappropiated by these communities, and have gained different meanings and connotations in African American English (AAE), yet these terms remain offensive when used by outsiders~\cite{sap2019risk}. Thus, annotators may be more likely to classify an AAE tweet as offensive when it is not, a risk that may be exacerbated if annotators are not themselves familiar with AAE~\cite{fazelpour2022diversity}. We use a recently published offensive language dataset collected by \citet{keswani2021towards}, which contains 1471 Twitter posts, and is a subset of the 25k Twitter post curated by \citet{davidson2017automated}. \citet{keswani2021towards} randomly assign the Twitter posts to 170 Amazon Mechanical Turk (Mtruk) labelers, so that each post is annotated by around 10 Mturk labelers regarding whether it contains offensive language. Thus, every post in \citet{keswani2021towards}'s dataset is associated with 1) a ``gold standard'' label that indicates whether it contain hate speech or offensive language based on ~\citet{davidson2017automated}'s dataset; 2) a dialect feature that indicates whether the tweet's dialect is AAE. 3) around 10 newly acquired labels from different Amazon Mechanical Truk (Mturk) labelers. 

\begin{figure}[h!]
\centering
  \includegraphics[scale = 0.4]{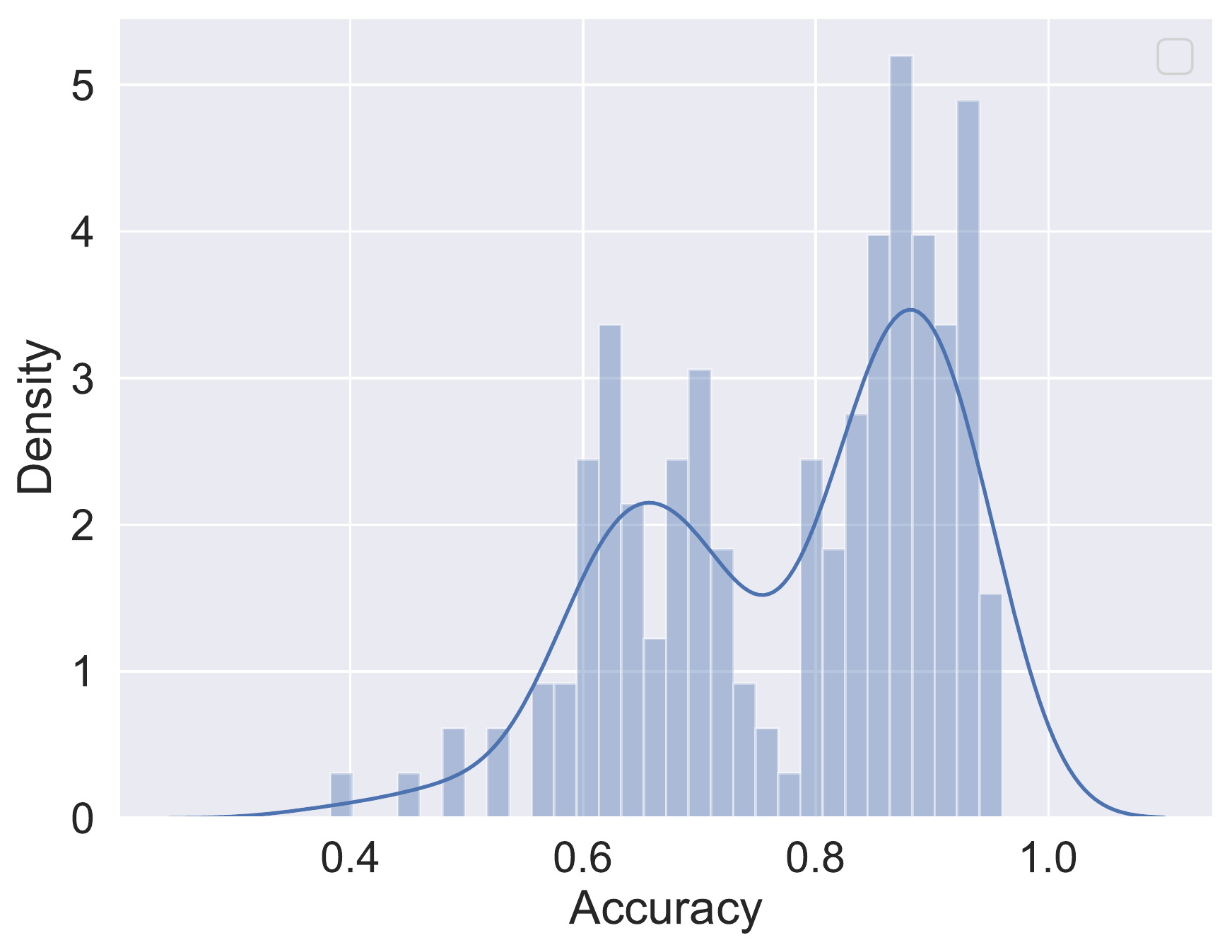}
\caption{Offensive language annotation accuracy distribution of the 170 crowd labelers, evaluated on the ``gold standard'' labels. There is a considerable discrepancy on crowd workers' performance of accurately flagging the tweets that contain hate speech or offensive language. }
\label{fig:acc_labeler}
\end{figure}

\begin{figure*}
\centering
\begin{minipage}[b]{0.5\textwidth}
\centering
\includegraphics[scale = 0.4]{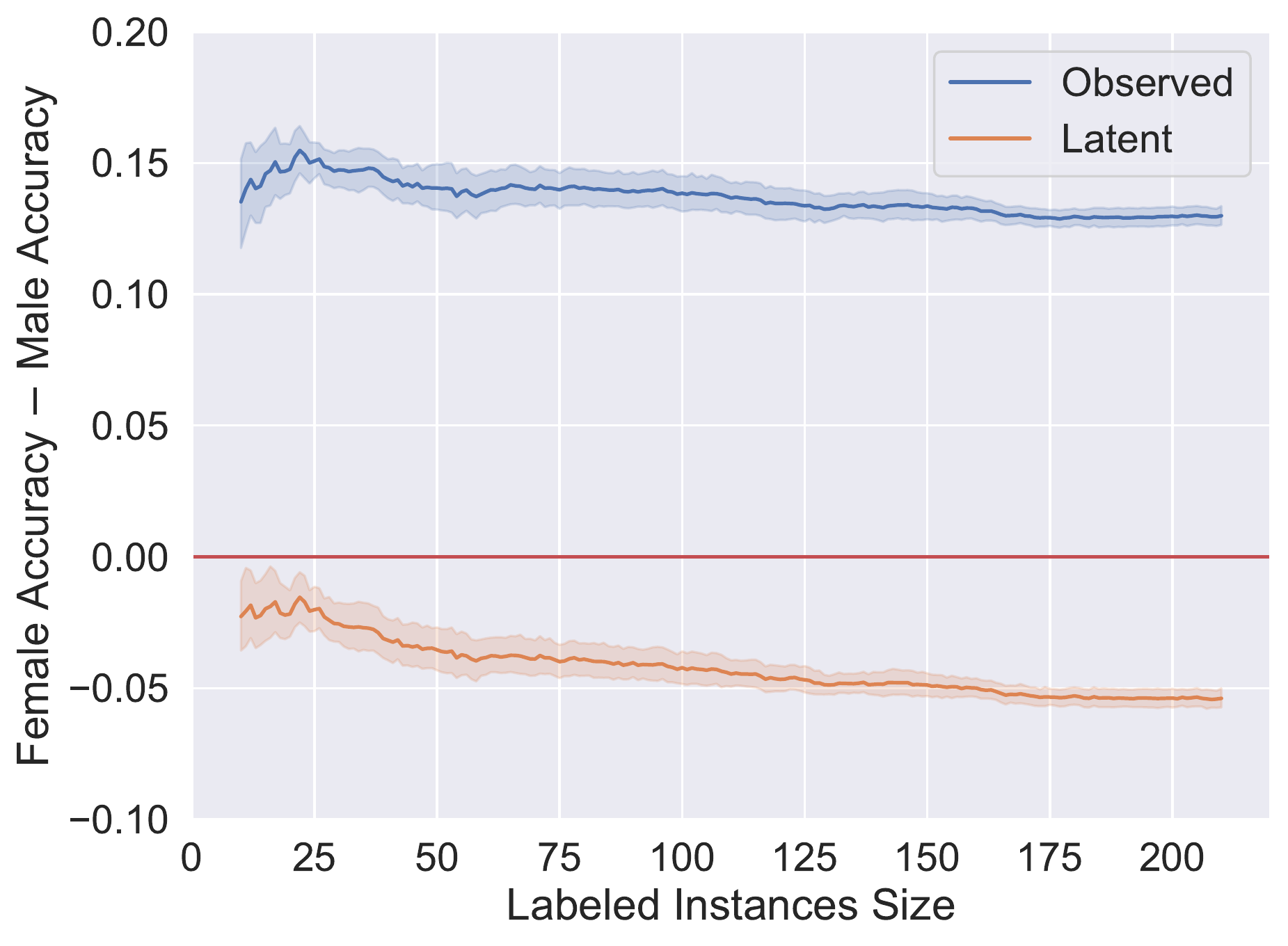}
\put(-178,158){(a)}
\end{minipage}%
\begin{minipage}[b]{0.5\textwidth}
\centering
\includegraphics[scale = 0.4]{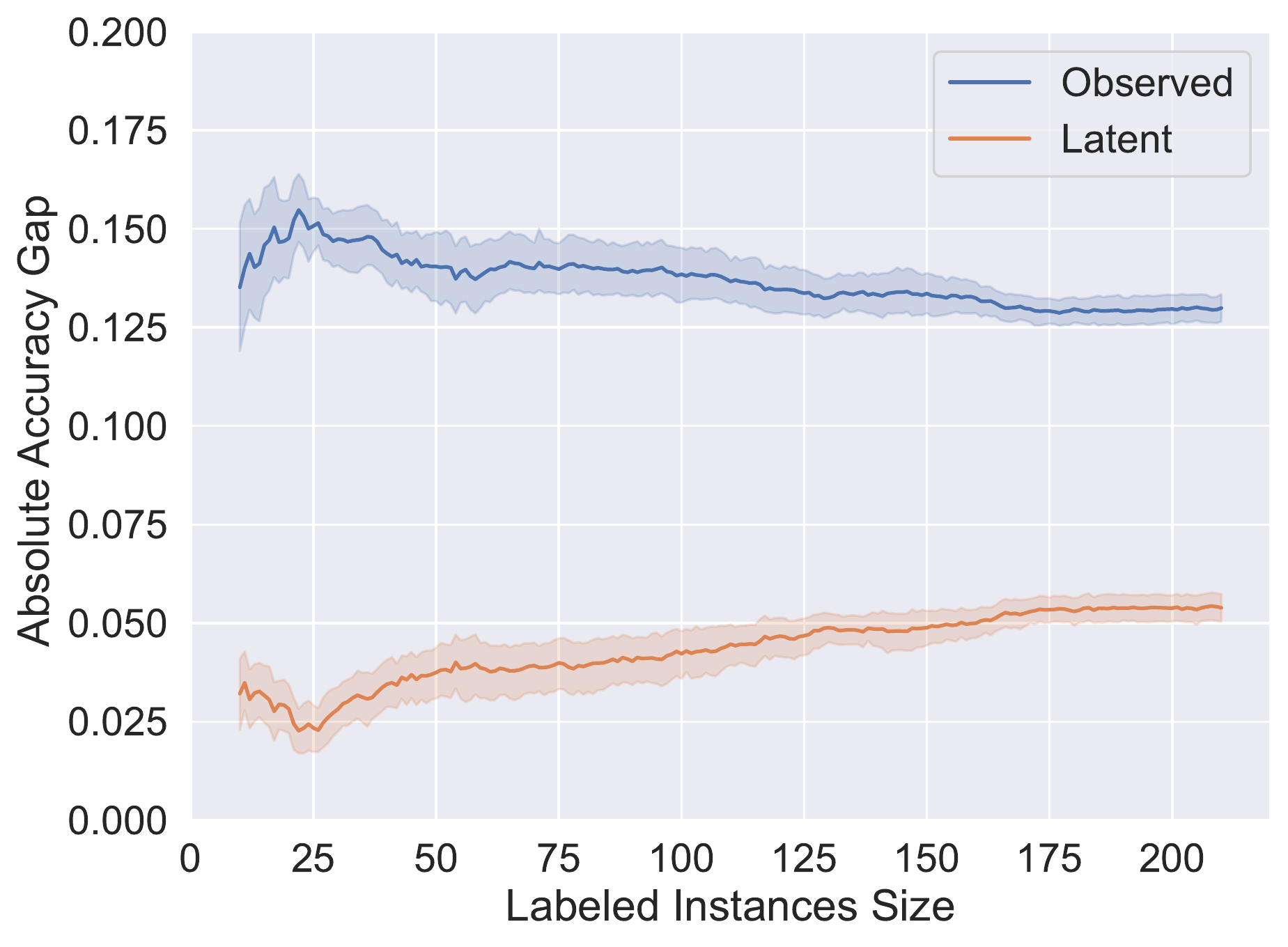}
\put(-178,158){(b)}
\end{minipage}%
\caption{Gender accuracy gap (left plot) and absolute gender accuracy gap (right plot) versus number of newly acquired training instances. Shaded areas indicate 95\% confidence interval. There are two insights: 1) misidentified disadvantaged group: evaluation on observed labels indicating male group to be the disadvantaged group, while in reality, female should be the disadvantaged group. 2) Bias may be interpreted as decreasing while the true bias is increasing.  }
\label{fig:AccGap_Adult}
\end{figure*}

Since the Mturk labelers were carefully recruited \cite{keswani2021towards}, it is reasonable to assume that the labelers have the intention to provide accurate labels despite being potentially affected by biases and knowledge limitations. And because different people may have different knowledge and carry different biases based on their background and personal experience, labelers' performance also varies. Figure \ref{fig:acc_labeler} shows the accuracy distribution of labelers assessed with respect to the original labels provided by~\citet{davidson2017automated}. Our goal is to assess how observed biased labels may affect learning when we use (fair) active learning to acquire new training instances. To do this, we curate a set of labels that correspond to a ``worst case scenario" in which a single labeler is available per instance, and the available label corresponds to the labeler with the largest bias against AAE tweets. Specifically, we deem the labels provided by \citet{davidson2017automated} as a gold standard, $Y^*$, and we use a subset of the labels collected by \citet{keswani2021towards} to construct $\tilde Y$. For each instance, we let $\tilde Y$ be the label acquired from the labeler who exhibits the largest performance disparity when evaluated with respect to $Y^*$. It is worth mentioning that the labels in \citet{davidson2017automated}'s dataset are majority votes of three or more crowd sourcing workers who were specifically instructed to label the tweet based on context, not the presence of particular words. Thus, we deem this as a gold standard because the instructions of the task are specifically designed to mitigate risks of bias stemming from ignoring context and over-relying on specific terms. 

\subsection{Experiment Setting}
We train active learning algorithms using the two datasets with their biased labels, to mimic the situation of using active learning to acquire more data labels while ignoring label bias. We then evaluate model performance using both observed biased labels and gold standard. Formally, let $\tilde Y \in \mathbb{R}^{n \times 1}$ and $Y^* \in \mathbb{R}^{n \times 1}$ be observed labels and gold standard, respectively. Let $X \in \mathbb{R}^{n \times d}$ be a matrix containing $d$ attributes for $n$ instances. We first split the observed dataset into training set $(X, \tilde Y)_\text{train}$ and testing set $(X,\tilde Y)_\text{test}$, and we use the training pool to perform active learning based on observed labels, $(X,\tilde Y)_\text{train}$.  Given a trained model $f$, we then obtain predictions $\hat Y = f(X_\text{test})$ on testing instances. Then, we evaluate $\hat Y_\text{test}$ with respect to both $\tilde Y_\text{test}$ and $Y^*_\text{test}$.
We perform the experiments on 10 random train and test partitions of the dataset (70-30 split), and consider the mean and confidence interval over the 10 random splits. For UCI Adult dataset, the maximum labeling budget was set at 200 (and we assume acquiring one label costs 1), after which the performance leveled out. Starting with 10 labeled instances (5 female and 5 male), we select one point to label at each active learning iteration in a consecutive order until the budget is depleted.
For the offensive language dataset, we start with 10 initial instances (5 AAE and 5 non-AAE) and  acquire 10 instances’ labels at each iteration. We specify the number of iteration to be 60 after which performance leveled off. We use the 100-dimensional vector representation of a tweet that obtained from GloVe \citep{pennington2014glove} pre-trained word embedding as input and predicts whether a tweet contains offensive language. 

For active learning strategies with fairness considerations, such as adaptive sampling, adaptive uncertainty sampling, and FAL algorithm, we need to choose a fairness measure that we interested in optimizing. We minimize true positive gap (TPR) gap between female and male groups for income level prediction on Adult dataset. as equalized opportunity~\cite{hardt2016equality} is a sensible fairness goal for similar prediction tasks in human resource domains. We minimize FPR gap between African American English (AAE) speaker and non-AAE speaker for offensive language detection task, as higher FPR for AAE tweets is the major inequality found in many offensive language detection algorithms \cite{sap2019risk}. \\

\section{Results and Analysis} \label{analysis}
In this section, we analyze the potential harms that can result from overlooking label bias in active label acquisition. To do this, we study the evolution of active learners (base models), 
comparing the \emph{observed} evaluation metrics and the evaluation metrics with respect to the gold standard. 
We first show findings when performing simulations using the Adult dataset, followed by the findings when using the real world offensive language dataset. 

We found three outstanding patterns in both scenarios. If we overlook label bias while acquiring labels: 1) collecting more data can lead to exacerbated bias; 2) data-driven strategies to identify the ``disadvantaged group" based on performance gaps can lead to misidentification; 3) relative comparisons of bias across models can be misleading, which may misguide model selection. 

\subsection{More Data Can Exacerbate Bias}\label{MoreDataAstray}
As we mentioned earlier in introduction and related work, calls to mitigate bias by collecting more data have gained attention in recent years. In particular, it is expected that fairness-aware data collection can help address this problem~\cite{chen2018my}. However, such strategies typically ignore potential label bias. We study the disparity between the bias mitigation performance evaluated on observed labels and the actual performance when evaluated on the gold standard labels. To accomplish so, we examine the accuracy (Acc) gaps and true positive rate (TPR) gaps, both in terms of the directed difference and the absolute gap. The directed accuracy gap which we used as one fairness metric is defined as the difference in accuracy between group $g_1$ and group $g_0$:

\begin{equation}
    \begin{split}
        \text{AccGap} = P(\hat Y = Y|G = g_0)\\
        -P(\hat Y = Y|G = g_1) 
    \end{split}
\end{equation}

Where $\hat Y$ and $Y$ are random variables representing predicted and testing labels, and $G$ is a random variable representing binary groups. $Y = \tilde Y$ when evaluated on observed labels and $Y = Y^*$ when evaluated on construct of interest or gold standard. As fairness metrics in proposed fairness-aware active learning algorithms often treat disparities as symmetric, we also visualize the accuracy gap as the absolute accuracy difference between $g_1$ and $g_0$:

\begin{equation}
    \begin{split}
        \text{Abs. AccGap} = | P(\hat Y = Y|G = g_0)\\
        -P(\hat Y = Y|G = g_1) |
    \end{split}
\end{equation}

Similarly, we examine both the directed TPR gap and absolute TPR gap between $g_1$ and $g_0$:

\begin{equation}
    \begin{split}
        \text{TPRGap} = P(\hat Y = Y|G = g_1, Y = 1)\\
        -P(\hat Y = Y|G = g0, Y = 1)\\
    \end{split}
\end{equation}
\begin{equation}
    \begin{split}
        \text{Abs. TPRGap} = |P(\hat Y = Y|G = g_1, Y = 1)\\
        -P(\hat Y = Y|G = g_0, Y = 1)|
    \end{split}
\end{equation}

In what follows, we present the results when applying the FAL algorithm for income prediction and applying uncertainty sampling for offensive language detection, which shows some of the most stark patterns of \emph{more data can exacerbate bias}. The results of the other four active learning algorithms assessed in the paper show a similar pattern and can be found in the Appendix.





\begin{figure*}
\centering
\begin{minipage}[b]{0.5\textwidth}
\centering\includegraphics[scale = 0.4]{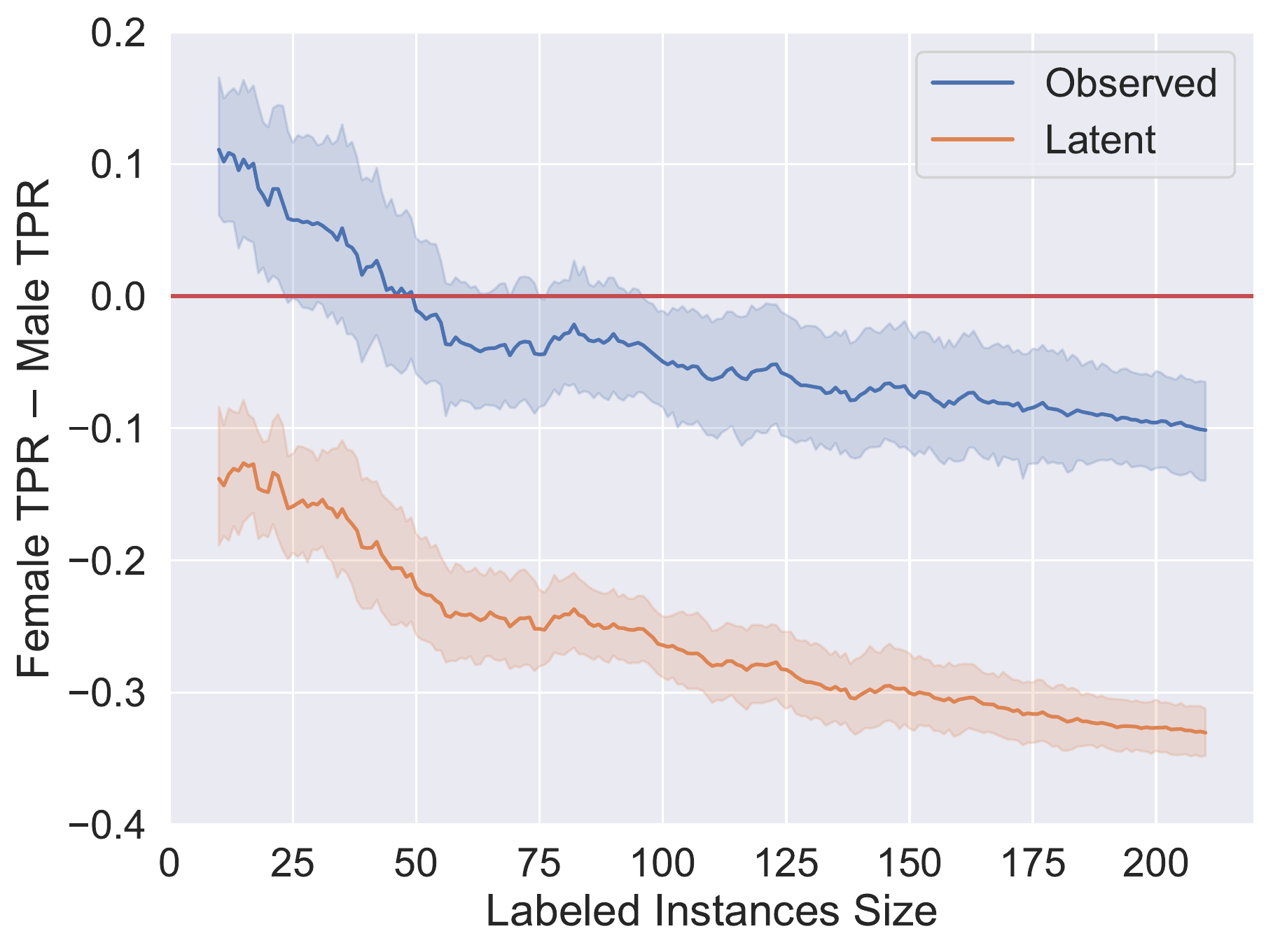} 
\put(-178,158){(a)}
\end{minipage}%
\begin{minipage}[b]{0.5\textwidth}
\centering\includegraphics[scale = 0.4]{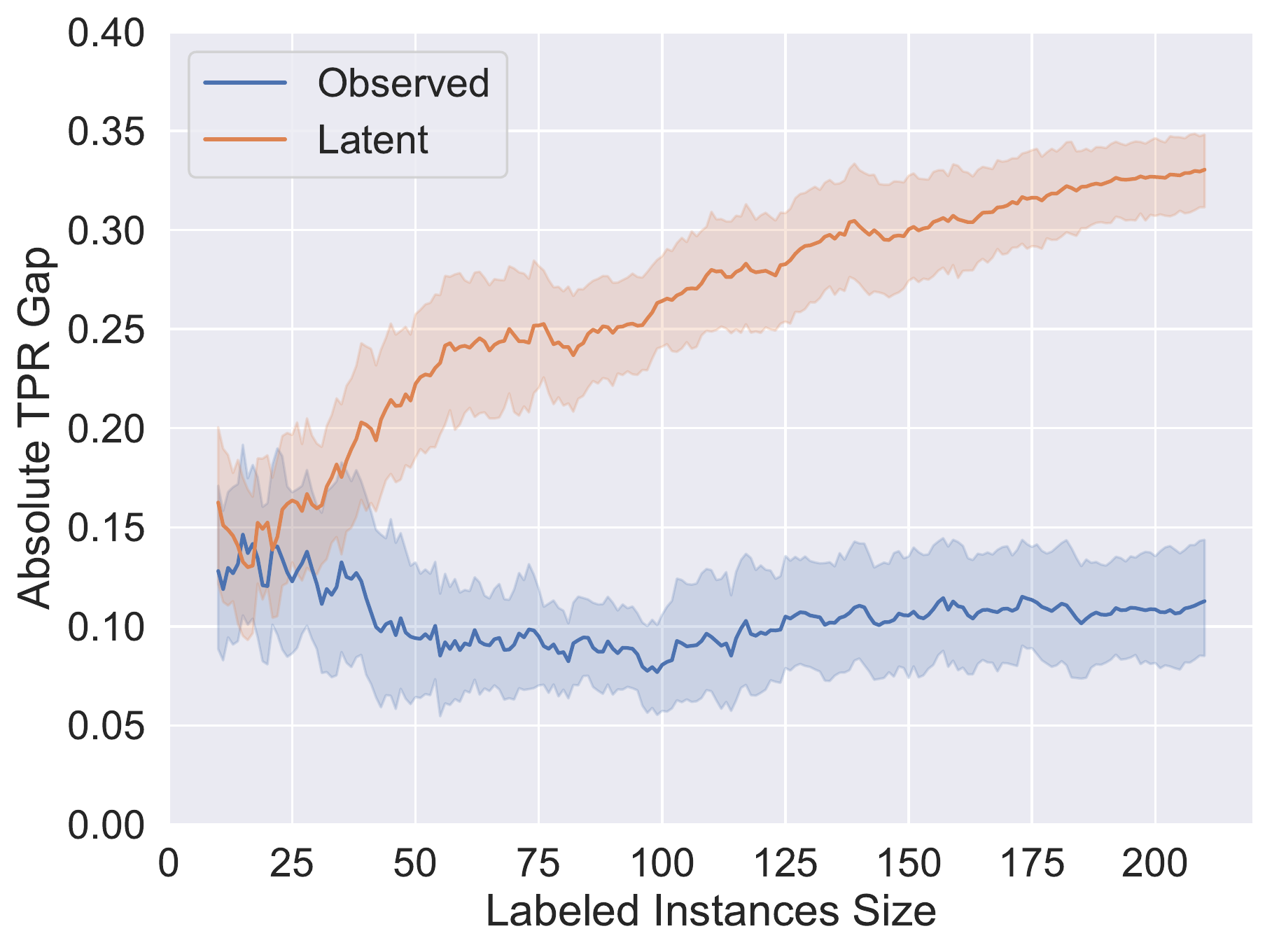}
\put(-178,158){(b)}
\end{minipage}%
\captionsetup{singlelinecheck =true, justification=justified}
\caption{Gender TRP gap (left plot) and absolute gender TPR gap (right plot) versus number of newly acquired taining instances. Shaded areas indicate 95\% confidence interval. Evaluation on observed labels shows data collection mitigates the gender TPR gap if we carefully choosing the stopping point, say, if we stop around 80 newly labeled instances. However, in reality, collecting more data further exacerbated the existing gender TPR gap. 
}
\label{fig:TprGap_Adult}
\end{figure*}

\subsubsection{Income Prediction} We begin by analyzing experiment results on the Adult dataset. 
Figure \ref{fig:AccGap_Adult} (a) shows the dynamics of female group's advantage in terms of accuracy as compared to male group (AccGap) during the data collection process under FAL algorithm, tested both on observed labels and on the (simulated) construct of interest. The x-axis indicates the number of newly acquired training instances using the FAL algorithm. The blue line represents AccGap tested on observed labels and the orange line represent AccGap tested on construct of interest. The shaded area depicts 95\% confidence interval of the AccGap. Note that an accuracy gap of zero, visualized by the red horizontal line, marks the position where male accuracy and female accuracy are perfectly equal to each other. According to figure \ref{fig:AccGap_Adult} (a), evaluation on observed labels shows that female accuracy is greater than male accuracy and data collection moderately mitigates this inequality. Meanwhile, evaluation on the construct of interest shows a very different pattern, indicating that female accuracy is lower than male accuracy, and more data leads to an enlarged accuracy gap and exacerbated the bias. 

Figure \ref{fig:AccGap_Adult} (b) illustrates the absolute accuracy gap. It is clear that the bias evolution trend is opposite when evaluated on the observed labels versus on the construct of interest. According to observed labels, the bias is reduced during the data collection process under the FAL framework, which aligns with the goal of collecting data to improve fairness. However, evaluation on the gold standard label indicates that the bias is actually aggravated as more data is collected. 

 Figure \ref{fig:TprGap_Adult} illustrates the TPRGap of the FAL algorithm during the data collection process, tested on observed labels (blue line) and construct of interest (orange line). According to figure \ref{fig:TprGap_Adult} (a), if label bias is overlooked, the evaluation on observed labels indicates female TPR started as greater than male TPR and the data collection mitigates the inequality and may over-correct the bias when newly collected instances size greater than 80. In such case, one may claim that the data collection is useful to mitigate bias. However, in reality, female TPR is initially less than male TPR and collecting more data exacerbated this inequality. Similarly, the visualization of the absolute TPRGap in figure \ref{fig:TprGap_Adult} (b) shows the anonymous and symmetric bias as TPRGap is slightly decreased and then remains stable based on observed labels, whereas the bias continues to aggravate in reality.

\subsubsection{Offensive Language Detection}

\begin{figure}[h!]
\centering
  \includegraphics[scale = 0.4]{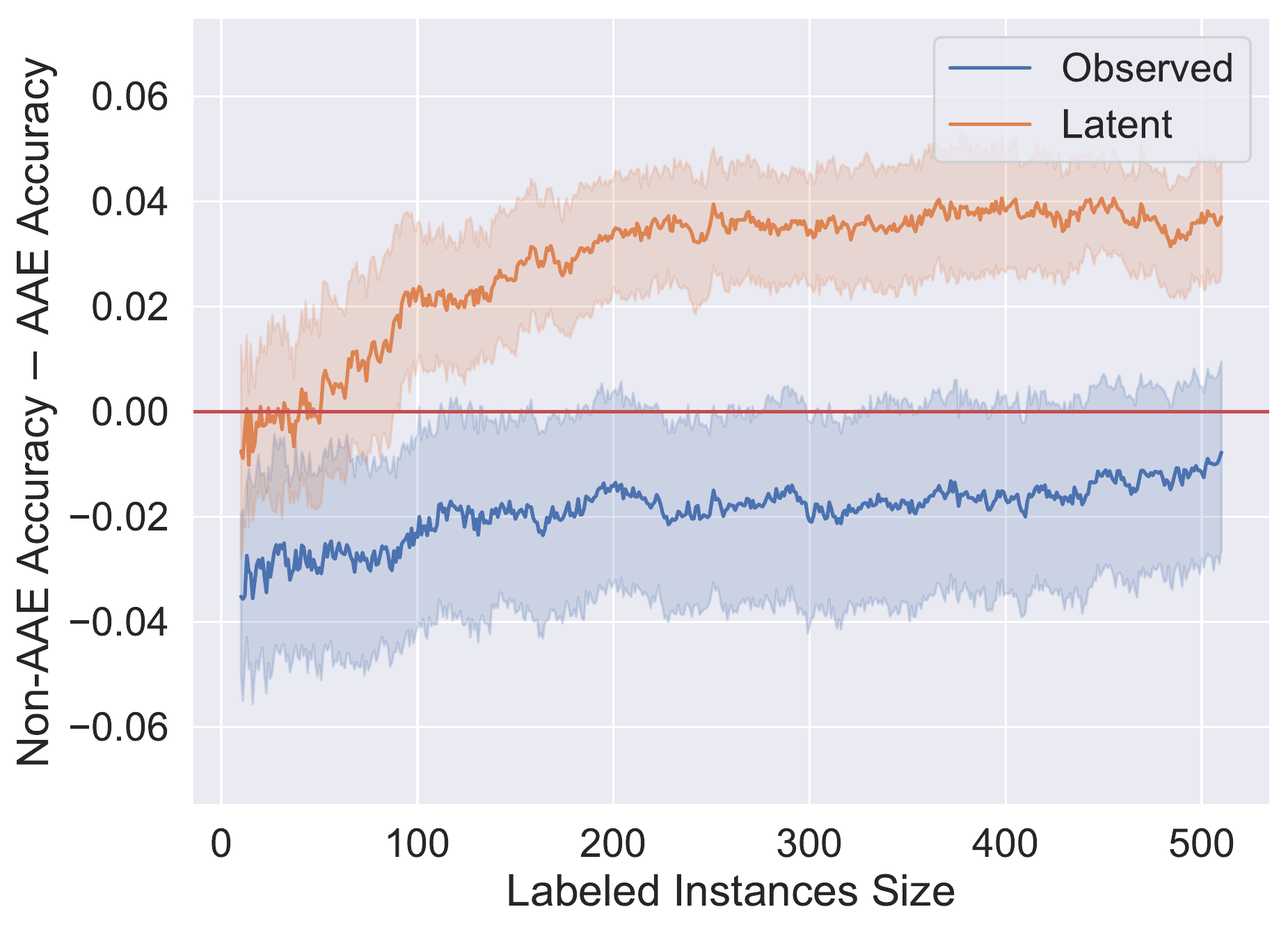}
\caption{Accuracy difference between non-AAE tweets and AAE tweets versus number of newly acquired training instance evaluated on observed labels (blue line) and gold standard (orange line). Shaded area indicates 95\% confidence interval. Evaluation on observed labels indicates non-AAE accuracy is less than AAE accuracy and data collection can mitigate the inequality. However, evaluation on the gold standard shows non-AAE accuracy is greater than AAE accuracy and data collection exacerbated this inequality. }
\label{fig:AccGap_Hate}
\end{figure}

We find a similar \emph{more data can exacerbate bias} pattern in offensive language detection experiments based on a real world dataset. As we described in the offensive language dataset section, racial bias is a central issue in offensive language detection tasks, considering the sensitive feature as indicating whether the dialect is African American English (AAE) or not. Let non-AAE be $g_0$ and AAE be $g_1$. 
Figure \ref{fig:AccGap_Hate} illustrates the dynamics of non-AAE group's advantage in terms of accuracy as compared to AAE group during the data collection process under uncertainty sampling.  It is clear that while the observed accuracy difference is approaching 0 as we acquire more data, the latent accuracy difference is getting larger. If we ignore the bias in the labels and depend just on the observed testing results to assess bias reduction performance, we might conclude that data collection aids bias mitigation, when in fact it exacerbates the problem. 


\subsection{Misidentifying Disadvantaged Group}
Many works on algorithmic fairness propose the idea of automatically identifying the disadvantaged group by evaluating a measure of interest for different groups, then mitigating the bias based on the evaluation. For example, adaptive sampling iteratively evaluates and identifies the disadvantaged group, then samples new instances from that group. Similarly, FAL relies on selecting the instances that are most helpful for mitigating the gap of a measure of interest, which implicitly identifies the disadvantaged group and tries to improve the model's performance on the group that exhibits the worse performance (according to some fairness measure) at a given point. 

Data-driven approaches to identify who the disadvantaged group is ignore contextual factors and historical inequities. This can be particularly problematic when such inequities result in label bias, in which case there is a risk of misidentifying who the disadvantaged group is, resulting in counterproductive mitigation strategies. 
Figures \ref{fig:AccGap_Adult}, \ref{fig:TprGap_Adult}, \ref{fig:AccGap_Hate} all demonstrate the risk of misidentifying the disadvantaged group. 

\subsubsection{Income Prediction}
In the context of income prediction, we know that women have historically less access to opportunity and high salaries, and have had lower income levels than men. The Adult dataset verifies the pattern as the ratio of males earning more than \$50K per year is three times higher than that of females in the dataset. In our experiments, it can be seen that if the observed label has a biased relationship with a latent construct of interest, such as the one we simulate, a reliance on the biased observed labels could lead females to be misidentified as the advantaged group, aggravating existing disparities.

\subsubsection{Offensive Language Detection}
When considering offensive language detection, researchers have found that ``AAE tweets and tweets
by self-identified African Americans are up to two times more likely to be labelled as offensive compared to others"~\cite{sap2019risk}. However, relying on evaluation on biased human labeling may result in treating AAE as the advantaged group, further marginalizing African Americans' voice. 

Most proposed bias mitigation algorithms are anonymous and symmetric, a property that is often lauded as a favorable trait, since application of such algorithms requires minimal contextual knowledge. However, with the prevalence of label bias, those algorithms run a risk of misidentifying the disadvantaged group, and thus  mitigation can become exacerbation. 
Many fairness-aware active learning algorithms, such as adaptive sampling, rely on identifying the disadvantaged group at each iteration of the data collection process in order to collect data from the disadvantaged group; thus, misidentifying the disadvantaged group can violate the original intention of the active data acquisition method to reduce disparities. Less obviously but importantly, algorithms such as FAL actively select instances with two goals: improving overall accuracy and reducing non-directional fairness violation. In this case, implicitly misidentifying the disadvantaged group would misguide the algorithm to select instances that can improve the advantaged group's performance.

\begin{figure}[h!]
\centering
  \includegraphics[scale = 0.4]{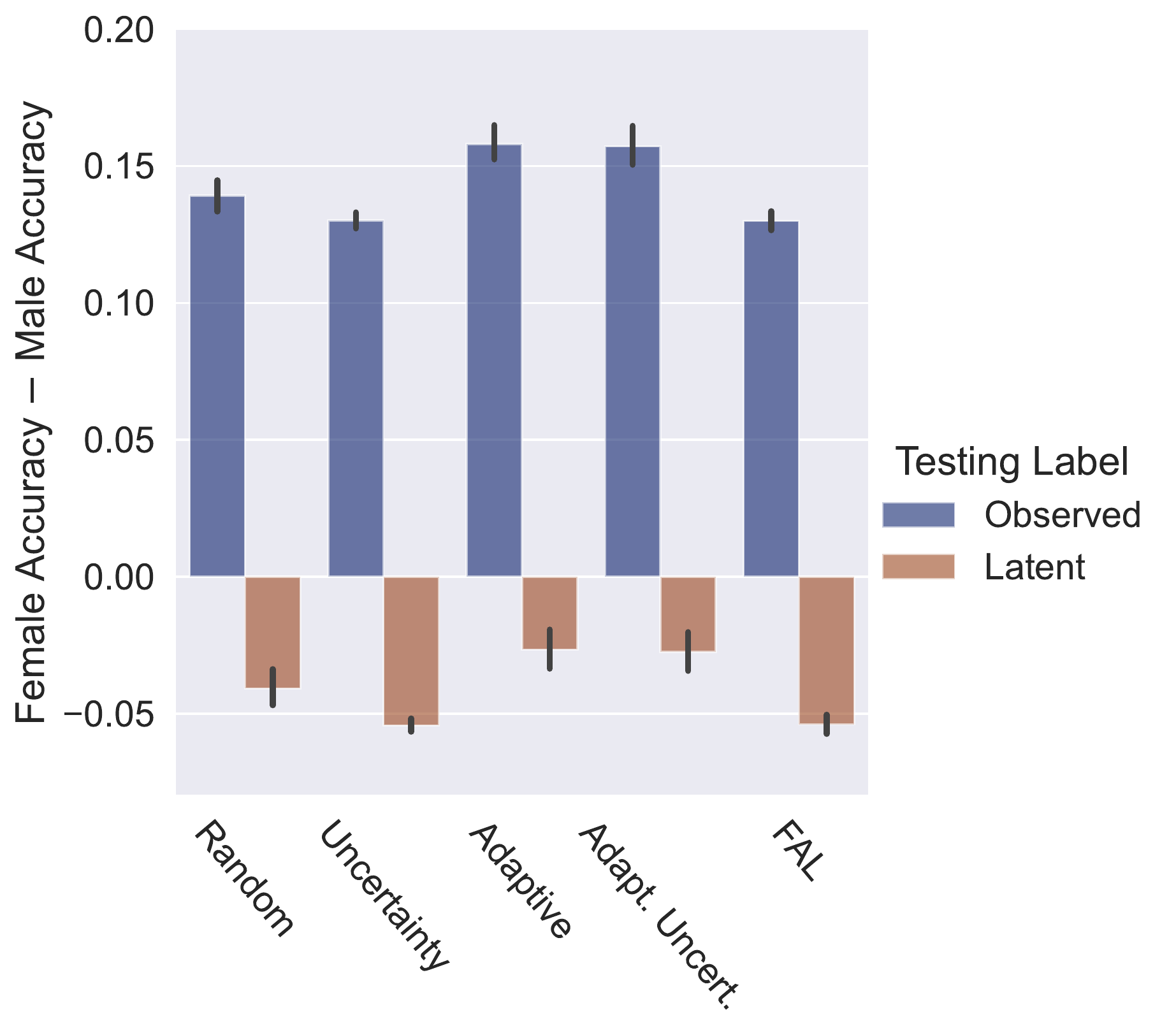}
\caption{Gender accuracy gap for all five active learning algorithms evaluated on observed labels (blue bars) and construct of interest (orange bars). The vertical line at middle of each bar illustrates 95\% confidence interval.  Model selection can be misguided: evaluation on observed labels indicates uncertainty sampling to be the most fair algorithm whereas it is the least fair algorithm in reality. }
\label{fig:AccDiff_adult}
\end{figure}

\subsection{Mislead Model Selection}


In addition to misguiding people's believes on bias mitigation performance of (fair) active learning assisted data collection, overlooking label bias can mislead model selection as well.  Suppose that multiple algorithms are considered, and fairness metrics are used to 
select the data acquisition strategy that provides the best bias mitigation effect from a list of algorithms, given comparable accuracy. 

\subsubsection{Income Prediction}
Figure \ref{fig:AccDiff_adult} illustrates accuracy difference between female and male groups in income prediction for all five active learning frameworks we described in `Methodology' section, evaluated on observed labels (blue bars) and gold standard (orange bars). 
By comparing the male accuracy minus female accuracy evaluated by biased observed labels (blue bars), we would select uncertainty sampling as the accuracy difference is the smallest among all strategies. However, the selected model, uncertainty sampling, could be the most biased model according to the evaluation on the construct of interest.  Therefore, if we overlook label bias, model selection can be misleading and compound existing bias. 

\subsubsection{Offensive Language Detection}
\begin{figure}[h!]
\centering
  \includegraphics[scale = 0.4]{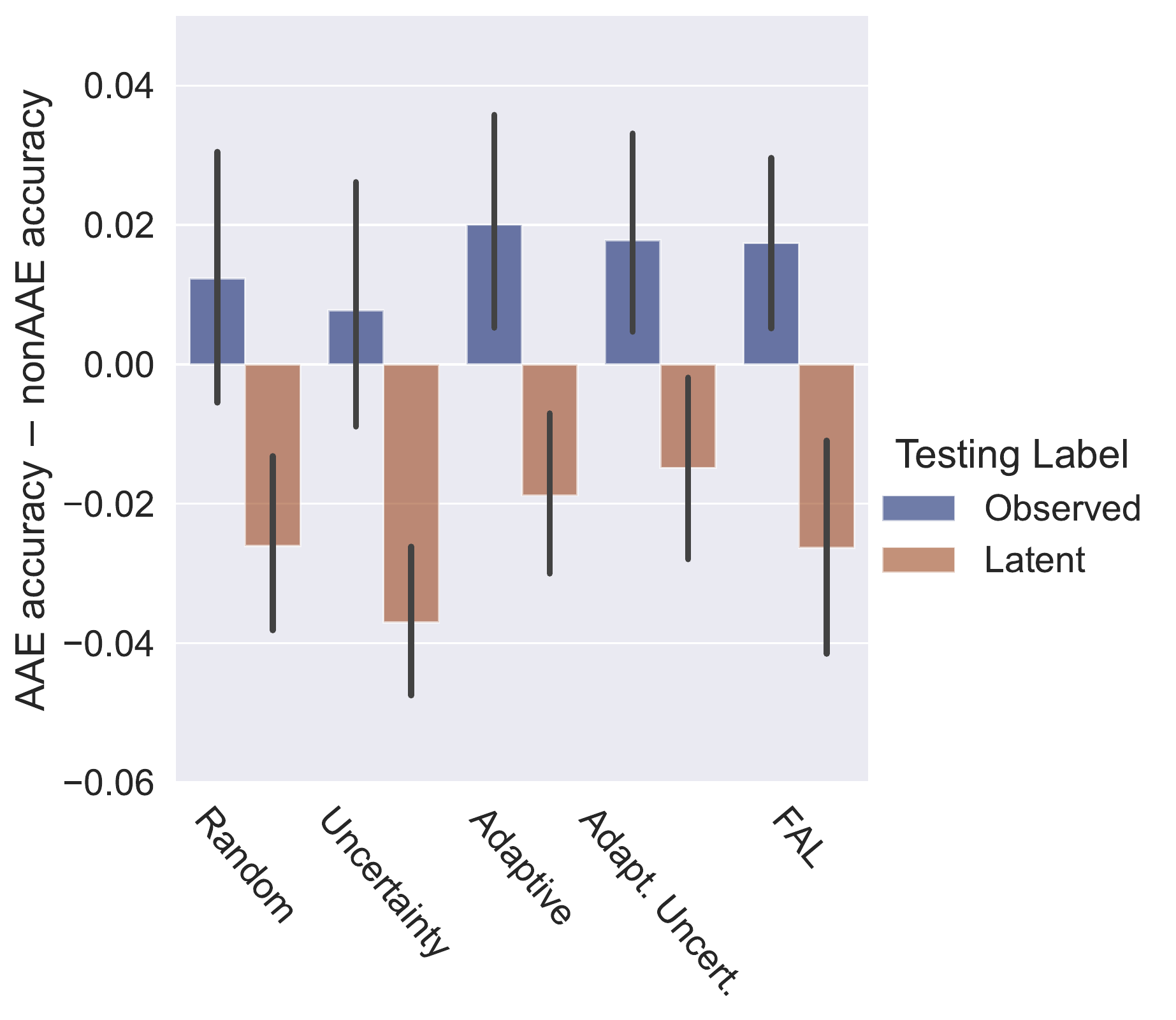}
\caption{Accuracy difference between AAE and non-AAE for all five active learning algorithms evaluated on observed labels (blue bars) and gold standard (orange bars). The vertical line at middle of each bar illustrates 95\% confidence interval.  Model selection can be misguided: evaluation on observed labels indicates uncertainty sampling to be the most fair algorithm whereas it is the least fair algorithm in reality.}
\label{fig:AccDiff_hate}
\end{figure}

The same pattern can be found in offensive language detection experiments as well. Figure \ref{fig:AccDiff_hate} illustrates accuracy difference between AAE and non-AAE tweets for all five active learning algorithms. Based on the evaluation metrics uncertainty sampling could be selected as the best algorithm for bias mitigation, while it may actually be the most biased algorithm with respect to the gold standard label.

\section{Conclusion}\label{conclusion}
As supervised learning algorithms are increasingly used for guiding decision making in various high-stakes domains, and recognizing its reliance on potentially biased sources of labels, there is a growing need to understand the potential harm of label bias on (algorithm assisted) data collection. In this paper, we presented an overview of different types of label bias in the context of supervised learning system and conducted empirical studies that uncover the effect of label bias on (fair) active learning algorithms. We evaluated bias mitigation performance of the most commonly used active learning algorithms and the recently proposed fairness-aware active learning strategies using a combination of simulations and real-world data. Our study has demonstrated that if we overlook label bias while acquiring labels: 1) collecting more data can lead to exacerbated bias; 2) Using data-driven strategies to identify the ``disadvantaged group" based on performance gaps can lead to misidentification;  and 3) relative comparisons of bias across models can be misleading, which may misguide model selection.


Most of the existing work on algorithmic fairness overlooks label bias, and research emphasizing the risk of label bias has been primarily conceptual, with only a few studies providing empirical evidence of risks. To the best of our knowledge, our work is the first to study how label bias may mislead active data collection, and how the introduction of fairness constraints that overlook label bias may fail to address the problem.

In addition to showing how data collection can be lead astray by label bias, our work has implications for the discussion regarding the identification of ``disadvantaged groups". A growing body of research has proposed methods that automatically identify disadvantaged groups, e.g.~\cite{abernethy2020adaptive}, and the ``symmetric" idea of bias is a common property of algorithmic fairness algorithms that consider any disparities in error rates to be indicative of bias. Our work provides empirical evidence showing that this can be problematic and may inadvertently exacerbate harms to already marginalized groups, since label bias may lead to a misidentification of the groups that need protection or are harmed by performance disparities.

When we say more data may lead to exacerbated bias, this does not mean that data collection efforts are inherently unproductive. Rather, our findings underscore the need for developing fairness-aware active learning algorithms that consider label bias. For ML practitioners, it is important to check the underling assumptions of proposed algorithms, differentiate the types of bias that fairness-aware algorithms aims to address, and scrutinize the data generating process to better understand the potential issues in the data. Our work facilitates more productive communication around label bias in supervised learning systems, as well as motivates more application-grounded strategies to mitigate them. 

\subsection{Limitations and Future Work}
Label bias is often highly context specific. The patterns we found in this study may only represents a portion of the possible consequences of ignoring label bias. We encourage future works that explore other possible harms in different application domains and under different assumptions of the relationship between observed labels and constructs of interest. In particular, in our experiments using the Adult dataset, which rely on simulations, we have made simplifying assumptions regarding the relationship between observed labels and gold standard labels. This by no means aims to provide a faithful estimation of a specific construct, such as economic contribution or skills. We use this as an example to empirically study the risks of label bias in a simplified setting. 

In the context of offensive language data, we rely on real data for both gold standard labels and observed labels, which introduces the challenge that both labels may be flawed. The gold standard is more reliable than what we assume to be the observed labels due to (1) the instructions of the task that we consider the gold standard, which specifically aim to mitigate risks of bias stemming from ignoring context and over-relying on specific terms; (2) our construction of the ``observed pool" as containing the assessments of those who exhibit the largest disparities. This is useful to empirically study how inadvertently relying on biased annotations could mislead the data collection process. However, this provides an assessment with respect to a point that may itself contain some bias, underestimating the magnitude of the problem.   

Additionally, the harms we study have been focused on a definition of label bias grounded on group fairness. Considering label bias in relation to other definitions of fairness, such as individual fairness, may shed lights on patterns and risks that are not visible through a group fairness lens. Similarly,
our 
empirical evaluation is limited to statistical disparities, and does not consider dynamic, long term harms~\cite{liu2018delayed}. In some cases data collection may involve the allocation of goods and burdens, and thus studying the dynamic effects may visibilize additional risks that are beyond the scope of this study. 

\section{Acknowledgement}
This research was supported by a Google AI Award for Inclusion Research, the Machine Learning Laboratory\footnote{https://ml.utexas.edu/}, and Good Systems\footnote{http://goodsystems.utexas.edu/}, a UT Austin Grand Challenge to develop responsible AI technologies.

\end{document}